\pdfoutput=1
\documentclass{article}
\usepackage[final]{corl_2018}
\usepackage{graphicx}
\usepackage{algorithm}
\usepackage{algorithmic}
\usepackage{bbm}
\usepackage{amsmath}
\usepackage{amssymb}
\usepackage{wrapfig}
\usepackage{subcaption}
\usepackage{caption}
\usepackage{booktabs}
\usepackage{bm}
\usepackage[symbol]{footmisc}

\newcommand{\ours}{MB-MPO }
\newcommand{\vspacesection}{\vspace{-0.1 in}}
\newcommand{\vspacesubsection}{\vspace{-0.05 in}}

\usepackage[per=slash]{siunitx}
\usepackage{comment}
\usepackage{ulem}

\definecolor{OliveGreen}{RGB}{0,200,25}
\newcommand{\red}[1]{\textcolor{red}{#1}}

\newcommand{\darkgreen}[1]{\textcolor{OliveGreen}{#1}}

\newcommand{\added}[1]{\blue{#1}}
\newcommand{\replaced}[2]{\red{\ifmmode \text{\sout{\ensuremath{#1}}} \else \sout{#1}\fi} \darkgreen{#2}}
\newcommand{\removed}[1]{\red{\ifmmode\text{\sout{\ensuremath{#1}}}\else\sout{#1}\fi}}

	\renewcommand{\added}[1]{#1}
 	\renewcommand{\replaced}[2]{#2}
	\renewcommand{\removed}[1]{}

\newcommand{\removedfootnote}[1]{\footnote{\removed{#1}}}
\newcommand{\removedsubsection}[1]{\subsection{\texorpdfstring{\removed{#1}}{#1}}}

\title{Model-Based Reinforcement Learning via \\ Meta-Policy Optimization}

\author{
  Ignasi Clavera\thanks{Equal contribution} \\
  UC Berkeley \\
  \texttt{iclavera@berkeley.edu} \\
  \And
    Jonas Rothfuss$^*$ \\
  KIT, UC Berkeley \\
  \texttt{jonas.rothfuss@kit.edu} \\
    \And
  John Schulman \\
  OpenAI \\
    \And
  Yasuhiro Fujita \\
  Preferred Networks \\
   \And
  Tamim Asfour \\
  Karlsruhe Inst. of Technology (KIT) \\
   \And
  Pieter Abbeel \\
  UC Berkeley, Covariant.AI \\
}

\begin{document}
\vspace{-10 pt}
\maketitle

\vspace{-20 pt}
\begin{abstract}
Model-based reinforcement learning approaches carry the promise of being data efficient. However, due to challenges in learning dynamics models that sufficiently match the real-world dynamics, they struggle to achieve the same asymptotic performance as model-free methods.  We propose Model-Based Meta-Policy-Optimization (MB-MPO), an approach that foregoes the strong reliance on accurate learned dynamics models. Using an ensemble of learned dynamic models, MB-MPO meta-learns a policy that can quickly adapt to any model in the ensemble with one policy gradient step. This steers the meta-policy towards internalizing consistent dynamics predictions among the ensemble while shifting the burden of behaving optimally w.r.t. the model discrepancies towards the adaptation step. Our experiments show that MB-MPO is more robust to model imperfections than previous model-based approaches. Finally, we demonstrate that our approach is able to match the asymptotic performance of model-free methods while requiring significantly less experience.
\end{abstract}



\vspacesection
\section{Introduction}
\vspace{-0.05 in}
\removed{Deep reinforcement learning (RL) methods have demonstrated remarkable success on a variety of tasks, including mastering classical games~\cite{silver2016go}, video games~\cite{mnih2015humanlevel}, and robotic control~\cite{Schulman2015,lillicrap2015ddpg,levine2016end}.} \added{Most of the recent success in reinforcement learning was achieved using model-free reinforcement learning algorithms \cite{Schulman2015,lillicrap2015ddpg,silver2016go}.} Model-free (MF) algorithms tend to achieve optimal performance, are generally applicable, and are easy to implement. However, this is achieved at the cost of being data intensive, which is exacerbated when combined with high-capacity function approximators like neural networks. Their high sample complexity presents a major barrier to their application to robotic control tasks, on which data gathering is expensive.

In contrast, model-based (MB) reinforcement learning methods are able to learn with significantly fewer samples by using a learned model of the environment dynamics against which policy optimization is performed. Learning dynamics models can be done in a sample efficient way since they are trained with standard supervised learning techniques, \replaced{and allows}{allowing} the use of off-policy data. However, accurate dynamics models can often be far more complex than good policies. For instance, pouring water into a cup can be achieved by a fairly simple policy while modeling the underlying dynamics of this task is highly complex. Hence, model-based methods have only been able to learn good policies on a much more limited set of problems, and even when good policies are learned, they typically saturate in performance at a level well below their model-free counterparts \citep{Deisenroth2013, Pong2018}.

Model-based approaches tend to rely on accurate (learned) dynamics models to solve a task. If the dynamics model is not sufficiently precise, the policy optimization is prone to overfit on the deficiencies of the model, leading to suboptimal behavior or even to catastrophic failures. 
This problem is known in the literature as model-bias \citep{deisenroth2011pilco}.
Previous work has tried to alleviate model-bias by characterizing the uncertainty of the models and learning a robust policy \citep{ deisenroth2011pilco,Rajeswaran2016EPOpt:Ensembles, zhou1996robust, lim2013rlrobustmdp,Kurutach2018}, often using ensembles to represent the posterior. This paper also uses ensembles, but very differently.

We propose Model-Based Meta-Policy-Optimization (MB-MPO), an orthogonal approach to previous model-based RL methods: while traditional model-based RL methods rely on the learned dynamics models to be sufficiently accurate to enable learning a policy that also succeeds in the real world, we forego reliance on such accuracy. We are able to do so by learning an ensemble of dynamics models and framing the policy optimization step as a meta-learning problem. Meta-learning, in the context of RL, aims to learn a policy that adapts fast to new tasks or environments~\citep{Finn2017, Duan2016RL2:Learning, wang2017learningtr, Mishra2018, sung2017learn2learnmc}. Using the models as learned simulators, MB-MPO learns a policy that can be quickly adapted to any of the fitted dynamics models with one gradient step.
This optimization objective steers the meta-policy towards internalizing the parts of the dynamics prediction that are consistent among the ensemble while shifting the burden of behaving optimally w.r.t discrepancies between models towards the adaptation step.
This way, the learned policy exhibits less model-bias without the need to behave conservatively. While much is shared with previous MB methods in terms of how trajectory samples are collected and the dynamic models are trained, the use of (and reliance on) learned dynamics models for the policy optimization is fundamentally different. 

In this paper we show that 1) model-based policy optimization can learn policies that match the asymptotic performance of model-free methods while being substantially more sample efficient, 2) \ours consistently outperforms previous model-based methods on challenging control tasks, 3) learning is still possible when the models are strongly biased. 
The low sample complexity of our method makes it applicable to real-world robotics. For instance, we are able learn an optimal policy in high-dimensional and complex quadrupedal locomotion within two hours of real-world data. Note that the amount of data required to learn such policy using model-free methods is 10$\times$ - 100$\times$ higher, and, to the best knowledge of the authors, no prior model-based method has been able to attain the model-free performance in such tasks.

\vspacesection
\section{Related Work}
\vspace{-0.05 in}
In this section, we discuss related work, including model-based RL and approaches that combine elements of model-based and model-free RL. Finally, we outline recent advances in the field of meta-learning.
 
\textbf{Model-Based Reinforcement Learning:
Addressing Model Inaccuracies.}
Impressive results with model-based RL have been obtained using simple linear models~\citep{bagnell2001autonomous, abbeel2006using, levine2014learning, levine2016end}.
However, like Bayesian models \citep{deisenroth2011pilco, koller2009localgps, kamthe2017derl}, their application is limited to low-dimensional domains. Our approach, which uses neural networks (NNs), is easily able to scale to complex high dimensional control problems.
NNs for model learning offer the potential to scale to higher dimensional problems with impressive sample complexity~\citep{Nagabandi2017a,Chua2018, punjani2015heli, wahlstrom2015pix2torq}. 
A major challenge when using high-capacity dynamics models is preventing policies from exploiting model inaccuracies. Several works approach this problem of model-bias by learning a distribution of models \cite{Depeweg2017, Rajeswaran2016EPOpt:Ensembles, Kurutach2018, Chua2018}, or by learning adaptive models \citep{clavera2018learn2adapt, fu2015oneshot, Gu2016}.
We incorporate the idea of reducing model-bias by learning an ensemble of models. However, we show that these techniques do not suffice in challenging domains, and demonstrate the necessity of meta-learning for improving asymptotic performance.

Past work has also tried to overcome model inaccuracies through the policy optimization process. Model Predictive Control (MPC) compensates for model imperfections by re-planning at each step~\citep{lenz2015deepmpc}, but it suffers from limited credit assignment and high computational cost. Robust policy optimization~\citep{Rajeswaran2016EPOpt:Ensembles, zhou1996robust, lim2013rlrobustmdp} looks for a policy that performs well across models; as a result policies tend to be over-conservative. In contrast, we show that MB-MPO learns a robust policy in the regions where the models agree, and an adaptive one where the models yield substantially different predictions.

\textbf{Model-Based + Model-Free Reinforcement Learning.}
Naturally, it is desirable to combine elements of model-based and model-free to attain high performance with low sample complexity. Attempts to combine them can be broadly categorized into three main approaches.
First, differentiable trajectory optimization methods propagate the gradients of the policy or value function through the learned dynamics model~\citep{Mishra2017, Heess2015LearningGradients} . However, the models are not explicitly trained to approximate first order derivatives, and, when backpropagating, they suffer from exploding and vanishing gradients~\citep{Kurutach2018}.
Second, model-assisted MF approaches use the dynamics models to augment the real environment data by imagining policy roll-outs~\cite{Sutton1991, Gu2016, WeberImagination-AugmentedLearning,Nagabandi2017a}. These methods still rely to a large degree on real-world data, which makes them impractical for real-world applications. Thanks to meta-learning, our approach could, if needed, adapt fast to the real-world with fewer samples.
Third, recent work fully decouples the MF module from the real environment by entirely using samples from the learned models \citep{Feinberg2018, Kurutach2018}. These methods, even though considering the model uncertainty, still rely on precise estimates of the dynamics to learn the policy. In contrast, we meta-learn a policy on an ensemble of models, which alleviates the strong reliance on precise models by training for adaption when the prediction uncertainty is high. \citet{Kurutach2018} can be viewed as an edge case of our algorithm when no adaptation is performed.

\textbf{Meta-Learning.} 
Our approach makes use of meta-learning to address model inaccuracies. Meta-learning algorithms aim to learn models that can adapt to new scenarios or tasks with few data points. Current meta-learning algorithms can be classified in three categories. One approach involves training a recurrent or memory-augmented network that ingests a training dataset and outputs the parameters of a learner model~\citep{schmidhuber1987srl,andrychowicz2016learntolearn}.
Another set of methods feeds the dataset followed by the test data into a recurrent model that outputs the predictions for the test inputs~\citep{Duan2016RL2:Learning, santoro2016one}. The last category embeds the structure of optimization problems into the meta-learning algorithm~\citep{Finn2017,husken2000fastlearning,ravi2016optimization}. These algorithms have been extended to the context of RL~\citep{Duan2016RL2:Learning, wang2017learningtr, sung2017learn2learnmc, Finn2017}. Our work builds upon MAML~\cite{Finn2017}. However, while in previous meta-learning methods each task is typically defined by a different reward function, each of our tasks is defined by the dynamics of different learned models.

\vspacesection
\section{Background}
\vspace{-0.05 in}
\vspacesubsection
\subsection{Model-based Reinforcement Learning}
A discrete-time finite Markov decision process (MDP) $\mathcal{M}$ is  defined by the tuple $(\mathcal{S}, \mathcal{A}, p, r, \gamma, p_0, H)$. 
Here, $\mathcal{S}$ is the set of states, $\mathcal{A}$ the action space, $p(\bf{s}_{t+1}|\bf{s}_t, \bf{a}_t)$ the transition distribution,
$r: \mathcal{S} \times \mathcal{A} \rightarrow \mathbb{R}$ is a reward function, $p_0: \mathcal{S} \to \mathbb{R}_+$ represents the initial state distribution, $\gamma$ the discount factor, and $H$ is the horizon of the process. We define the return as the sum of rewards $r(\bm{s}_t, \bm{a}_t)$ along a trajectory $\tau := (\bm{s}_{0}, \bm{a}_{0}, ..., \bm{s}_{H-1}, \bm{a}_{H-1}, \bm{s}_{H})$. The goal of reinforcement learning is to find a policy $\pi: \mathcal{S} \times \mathcal{A} \rightarrow \mathbb{R}^+$ that maximizes the expected return. 

While model-free RL does not explicitly model state transitions, model-based RL methods learn the transition distribution, also known as dynamics model, from the observed transitions. This can be done with a parametric function approximator $\hat p_{\bm{\phi}}({\bf s'}|{\bf s}, {\bf a})$. In such case, the parameters $\bm{\phi}$ of the dynamics model are optimized to maximize the log-likelihood of the state transition distribution.

\vspacesubsection
\subsection{Meta-Learning for Reinforcement Learning}\label{sec:meta-rl}
Meta-RL aims to learn a learning algorithm which is able to quickly learn optimal policies in MDPs $\mathcal{M}_k$ drawn from a distribution $\rho(\mathcal{M})$ over a set of MDPs. The MDPs $\mathcal{M}_k$ may differ in their reward function $r_k(\bm{s}, \bm{a})$ and transition distribution $p_k(\bm{s}_{t+1}|\bm{s}_t, \bm{a}_t)$, but share action space $\mathcal{A}$ and state space $\mathcal{S}$.

Our approach builds on the gradient-based meta-learning framework MAML \cite{Finn2017}, which in the RL setting, trains a parametric policy $\pi_{\bm{\theta}}(\bm{a}|\bm{s})$ to quickly improve its performance on a new task with one or a few vanilla policy gradient steps. The meta-training objective for MAML can be written as:
\vspace{-0.1 in}
\begin{align}
\max_{\bm{\theta}} \enspace \mathbb{E}_{\begin{subarray}{l}{\mathcal{M}_k \sim \rho({\mathcal{M}})} \\
\bm{s}_{t+1} \sim p_k \\
    \bm{a}_t \sim \pi_{\bm{\theta'}}(\bm{a}_t | \bm{s}_t)\end{subarray}} \bigg[ \sum_{t=0}^{H-1} r_k(\bm{s}_t, \bm{a}_t) \bigg] \quad
\text{s.t.:} \enspace \bm{\theta}' = \bm{\theta} +  \alpha ~ \nabla_{ \bm{\theta}} \mathbb{E}_{\begin{subarray}{l}{
\bm{s}_{t+1} \sim p_k} \\
    \bm{a}_t \sim \pi_{\bm{\theta}}(\bm{a}_t | \bm{s}_t)\end{subarray}} \bigg[ \sum_{t=0}^{H-1} r_k(\bm{s}_t, \bm{a}_t)\bigg]
\end{align}
\vspace{-0.1 in}

MAML attempts to learn an initialization $\bm{\theta}^*$ such that for any task $\mathcal{M}_k \sim \rho(\mathcal{M})$ the policy attains maximum performance in the respective task after one policy gradient step.

\vspacesection
\section{Model-Based Meta-Policy-Optimization}
\vspace{-0.05 in}
Enabling complex and high-dimensional real robotics tasks requires extending current model-based methods to the capabilities of mode-free while, at the same time, maintaining their data efficiency. Our approach, model-based meta-policy-optimization (MB-MPO), attains such goal by framing model-based RL as meta-learning a policy on a distribution of dynamic models, advocating to maximize the policy adaptation, instead of robustness, when models disagree. This not only removes the arduous task of optimizing for a single policy that performs well across differing dynamic models, but also results in better exploration properties and higher diversity of the collected samples, which leads to improved dynamic estimates.

We instantiate this general framework by employing an ensemble of learned dynamic models and meta-learning a policy that can be quickly adapted to any of the dynamic models with one policy gradient step. In the following, we first describe how the models are learned, then explain how the policy can be meta-trained on an ensemble of models, and, finally, we present our overall algorithm.

\vspacesubsection
\subsection{Model Learning}\label{sec:model-learning}
A key component of our method is learning a distribution of dynamics models, in the form of an ensemble, of the real environment dynamics. In order to decorrelate the models, each model differs in its random initialization and it is trained with a different randomly selected subset $\mathcal{D}_k$ of the collected real environment samples. In order to address the distributional shift that occurs as the policy changes throughout the meta-optimization, we frequently collect samples under the current policy, aggregate them with the previous data $\mathcal{D}$, and retrain the dynamic models with warm starts.

In our experiments, we consider the dynamics models to be a deterministic function of the current state $\bm{s}_t$ and action $\bm{a}_t$, employing a feed-forward neural network to approximate them. We follow the standard practice in model-based RL of training the neural network to predict the change in state $\Delta \bm{s} = \bm{s}_{t+1} - \bm{s}_t$ (rather than the next state $\bm{s}_{t+1}$)~\citep{Nagabandi2017a,deisenroth2011pilco}. We denote by $\hat{f}_{\bm{\phi}}$ the function approximator for the next state, which is the sum of the input state and the output of the neural network.
The objective for learning each model $\hat{f}_{\bm{\phi}_k}$ of the ensemble is to find the parameter vector $\bm{\phi}_k$ that minimizes the $\ell_2$ one-step prediction loss:
\begin{equation}
\min_{\bm{\phi}_k} \frac{1}{|\mathcal{D}_k|}\sum_{(\bm{s}_t,\bm{a}_t,\bm{s}_{t+1})\in \mathcal{D}_k}\|\bm{s}_{t+1}-\hat{f}_{\bm{\phi}_k}(\bm{s}_t,\bm{a}_t)\|_2^2
\label{eq:model-learning}
\end{equation}
where $\mathcal{D}_k$ is a sampled subset of the training data-set $\mathcal{D}$ that stores the transitions which the agent has experienced. Standard techniques to avoid overfitting and facilitate fast learning are followed; specifically, 1) early stopping the training based on the validation loss, 2) normalizing the inputs and outputs of the neural network, and 3) weight normalization \citep{Salimans2016}.

\vspacesubsection
\subsection{Meta-Reinforcement Learning on Learned Models} \label{sec:meta-training}
Given an ensemble of learned dynamic models for a particular environment, our core idea is to learn a policy which can adapt quickly to any of these models. To learn this policy, we use gradient based meta-learning with MAML (described in Section~\ref{sec:meta-rl}). To properly formulate this problem in the context of meta-learning, we first need to define an appropriate task distribution.
Considering the models $\{ \hat{f}_{\bm{\phi}_1}, \hat{f}_{\bm{\phi}_2}, ..., \hat{f}_{\bm{\phi}_K} \}$, which approximate the dynamics of the true environment, we can construct a uniform task distribution by embedding them into different MDPs $\mathcal{M}_k=(S, A, \hat{f}_{\bm{\phi}_k}, r, \gamma, p_0)$ using these learned dynamics models. We note that, unlike the experimental considerations of prior methods~\cite{Duan2016RL2:Learning,Finn2017,Mishra2018}, in our work the reward function remains the same across tasks while the dynamics vary. Therefore, each task constitutes a different belief about what the dynamics in the true environment could be. Finally, we pose our objective as the following meta-optimization problem:
\begin{align}
\max_{\bm{\theta}} \quad \frac{1}{K} \sum_{k=0}^K J_k(\bm{\theta'}_k) \quad \quad  \text{s.t.:}  \quad \bm{\theta'}_k = \bm{\theta} + \alpha ~ \nabla_{\bm{\theta}} J_k(\bm{\theta}) \label{eq:meta-objective}
\end{align}
with $J_k(\bm{\theta})$ being the expected return under the policy $\pi_{\bm{\theta}}$ and the estimated dynamics model $\hat f_{\bm{\phi}_k}$.
\begin{align}
J_k(\bm{\theta}) = \mathbb{E}_{\bm{a}_t \sim \pi_{\bm{\theta}}(\bm{a}_t | \bm{s}_t)} \bigg[ \sum_{t=0}^{H-1} r(\bm{s}_t, \bm{a}_t) \bigg| \bm{s}_{t+1} = \hat f_{\bm{\phi}_k}(\bm{s}_t, \bm{a}_t) \bigg]
\label{reward_objective}
\end{align}
For estimating the expectation in Eq.~\ref{reward_objective} and computing the corresponding gradients, we sample trajectories from the imagined MDPs. The rewards are computed by evaluating the reward function, which we assume as given, in the predicted states and actions $r(\hat{f}_{\bm{\phi}_k}(\bm{s}_{t-1}, \bm{a}_{t-1}, \bm{a}_t))$.
In particular, when estimating the adaptation objectives $J_k(\bm{\theta})$, the meta-policy $\pi_{\bm{\theta}}$ is used to sample a set of imaginary trajectories $\mathcal{T}_k$ for each model $\hat{f}_{\bm{\phi}_k}$. 
For the meta-objective $\frac{1}{K} \sum_{k=0}^K J_k(\bm{\theta'}_k)$, we generate trajectory roll-outs $\mathcal{T}'_k$ with the models $\hat{f}_{\bm{\phi}_k}$ and the policies $\pi_{\bm{\theta'}_k}$ obtained from adapting the parameters $\bm{\theta}$ to the $k$-th model. 
Thus, no real-world data is used for the data intensive step of meta-policy optimization.

In practice, any policy gradient algorithm can be chosen to perform the meta-update of the policy parameters. In our implementation, we use Trust-Region Policy Optimization (TPRO)~\citep{Schulman2015} for maximizing the meta-objective, and employ vanilla policy gradient (VPG)~\cite{peters2006pgm} for the adaptation step. To reduce the variance of the policy gradient estimates a linear reward  baseline is used.

\vspacesubsection
\subsection{Algorithm}
In the following, we describe the overall algorithm of our approach (see Algorithm~\ref{alg1}).
First, we initialize the models and the policy with different random weights. Then, we proceed to the data collection step. In the first iteration, a uniform random controller is used to collect data from the real-world, which is stored in a buffer $\mathcal{D}$. At subsequent iterations, trajectories from the real-world are collected with the adapted policies $\{ \pi_{\bm{\theta'}_1},...,\pi_{\bm{\theta'}_K} \}$, and then aggregated with the trajectories from previous iterations. 
The models are trained with the aggregated real-environment samples following the procedure explained in section~\ref{sec:model-learning}.
The algorithm proceeds by imagining trajectories from each the ensemble of models $\{f_{\bm{\phi}_1}, ..., f_{\bm{\phi}_K} \}$ using the policy $\pi_{\bm{\theta}}$. 
These trajectories are are used to perform the inner adaptation policy gradient step, yielding the adapted policies $\{ \pi_{\bm{\theta'}_1},...,\pi_{\bm{\theta'}_K} \}$. 
Finally, we generate imaginary trajectories using the adapted policies $\pi_{\bm{\theta'}_k}$ and models $f_{\bm{\phi}_k}$, and optimize the policy towards the meta-objective (as explained in section~\ref{sec:meta-training}). 
We iterate through these steps until desired performance is reached. The algorithm returns the optimal pre-update parameters $\bm{\theta^*}$.

\begin{algorithm}[t] 
\caption{\ours}
\label{alg1} 
\begin{algorithmic}[1] 
	\REQUIRE Inner and outer step size $\alpha$, $\beta$
    \STATE Initialize the policy $\pi_{\bm{\theta}}$, the models $\hat{f}_{\bm{\phi}_1}, \hat{f}_{\bm{\phi}_2}, ..., \hat{f}_{\bm{\phi}_K}$ and $\mathcal{D} \leftarrow \emptyset$
    \REPEAT
    	 \STATE Sample trajectories from the real environment with the adapted policies $\pi_{\bm{\theta}_1^{'}}, ..., \pi_{\bm{\theta}_K^{'}}$. Add them to $\mathcal{D}$.
    	 \STATE Train all models using $\mathcal{D}$.
         \FORALL{models $\hat{f}_{\bm{\phi}_k}$}
            \STATE Sample imaginary trajectories $\mathcal{T}_k$ from $\hat{f}_{\bm{\phi}_k}$ using $\pi_{\bm{\theta}}$
            \STATE Compute adapted parameters $\bm{\theta'}_k = \bm{\theta} + \alpha ~\nabla_{\bm{\theta}}J_{k}(\bm{\theta})$ using trajectories $\mathcal{T}_k$ \label{algo1:inner_update}
            \STATE Sample imaginary trajectories  $\mathcal{T}'_k$ from $\hat{f}_{\bm{\phi}_k}$ using the adapted policy $\pi_{\bm{\theta'}_k}$
         \ENDFOR  
	\STATE Update $\bm{\theta} \rightarrow \bm{\theta} - \beta~\frac{1}{K}\sum_k \nabla_{\bm{\theta}} J_k(\bm{\theta'}_k)$ using the trajectories $\mathcal{T}'_k$  \label{algo1:outer_update}
    \UNTIL{the policy performs well in the real environment}
    \RETURN Optimal pre-update parameters $\bm{\theta^*}$
\end{algorithmic}
\end{algorithm}

\section{Benefits of the Algorithm} 
\label{section:benefits}

Meta-learning a policy over an ensemble of dynamic models using imaginary trajectory roll-outs provides several benefits over traditional model-based and model-based model-free approaches. \added{In the following we discuss several such advantages, aiming to provide intuition for the algorithm.}

\textbf{Regularization effect during training.} Optimizing the policy to adapt within one policy gradient step to any of the fitted models imposes a regularizing effect on the policy learning (as~\cite{NicholOnAlgorithms} observed in the supervised learning case). The meta-optimization problem steers the policy towards higher plasticity in regions with high dynamics model uncertainty, shifting the burden of adapting to model discrepancies towards the inner policy gradient update.

We consider plasticity as the policy's ability to change its (conditional) distribution with a small change (i.e. gradient update) in the parameter space. The policy plasticity is manifested in the statistical distance between the pre- and post-update policy. In section~\ref{section:plasticity} we analyze the connection between model uncertainty and the policy plasticity, finding a strong positive correlation between the model ensembles predictive variance and the KL-divergence between $\pi_{\bm{\theta}}$ and $\pi_{\bm{\theta'}_k}$.
This effect prevents the policy to learn sub-optimal behaviors that arise in robust policy optimization. More importantly, this regularization effect fades away once the dynamics models get more accurate, which leads to asymptotic optimal policies if enough data is provided to the learned models. In section~\ref{exp:robustness}, we show how this property allows us to learn from noisy and highly biased models.

\textbf{Tailored data collection for fast model improvement.} Since we sample real-environment trajectories using the different policies $\{ \pi_{\bm{\theta'}_1},...,\pi_{\bm{\theta'}_K} \}$ obtained by adaptation to each model, the collected training data is more diverse which promotes robustness of the dynamic models. Specifically, the adapted policies tend to exploit the characteristic deficiencies of the respective dynamic models. As a result, we collect real-world data in regions where the dynamic models insufficiently approximate the true dynamics. This effect accelerates correcting the imprecision of the models leading to faster improvement. \added{In Appendix ~\ref{appendix:taylored_data_colection}, we experimentally show the positive effect of tailored data collection on the performance}.

\textbf{Fast fine-tuning.} \added{Meta-learning optimizes a policy for fast adaptation~\cite{Finn2017} to a set of tasks. In our case, each task corresponds to a different believe of what the real environment dynamics might be. } When optimal performance is not achieved, the ensemble of models will present high discrepancy in their predictions, \added{increasing the likelihood of the real dynamics to lie in the believe distribution's support.} \removed{Consequently, the meta-optimization is likely to learn a policy that exhibits high adaptability towards different dynamics.}
As a result, \added{the learned policy is likely to exhibit high adaptability towards the real environment, and} fine-tuning the policy with VPG on the real environment leads to faster convergence than training the policy from scratch or from any other MB initialization.

\textbf{Simplicity.} Our approach, contrary to previous methods, is simple: it does not rely on parameter noise exploration, careful reinitialization of the model weights or policy's entropy, hard to train probabilistic models, and it does not need to address the model distribution mismatch~\cite{Chua2018, Kurutach2018, Feinberg2018}.

\vspacesection
\section{Experiments} \label{sec:experiments}
\vspace{-0.05 in}
The aim of our experimental evaluation is to examine the following questions: 
1) How does \ours compare against state-of-the-art model-free and model-based methods in terms of sample complexity and asymptotic performance? 
2) How does the model uncertainty influence the policy's plasticity?
3) How robust is our method against imperfect models? 

To answer the posed questions, we evaluate our approach on six continuous control benchmark tasks in the Mujoco simulator \cite{mujoco}. A depiction of the environments as well a detailed description of the experimental setup can be found in Appendix \ref{appendix:experiment_setup}.
In all of the following experiments, the pre-update policy is used to report the average returns obtained with our method. The performance reported are averages over at least three random seeds. The source code and the experiments data is available on our supplementary website \footnote{https://sites.google.com/view/mb-mpo}.

\vspacesubsection
\subsection{Comparison to State-of-the-Art: Model-Free}
\begin{figure}[h]
\centering
\includegraphics[width=.95\textwidth]{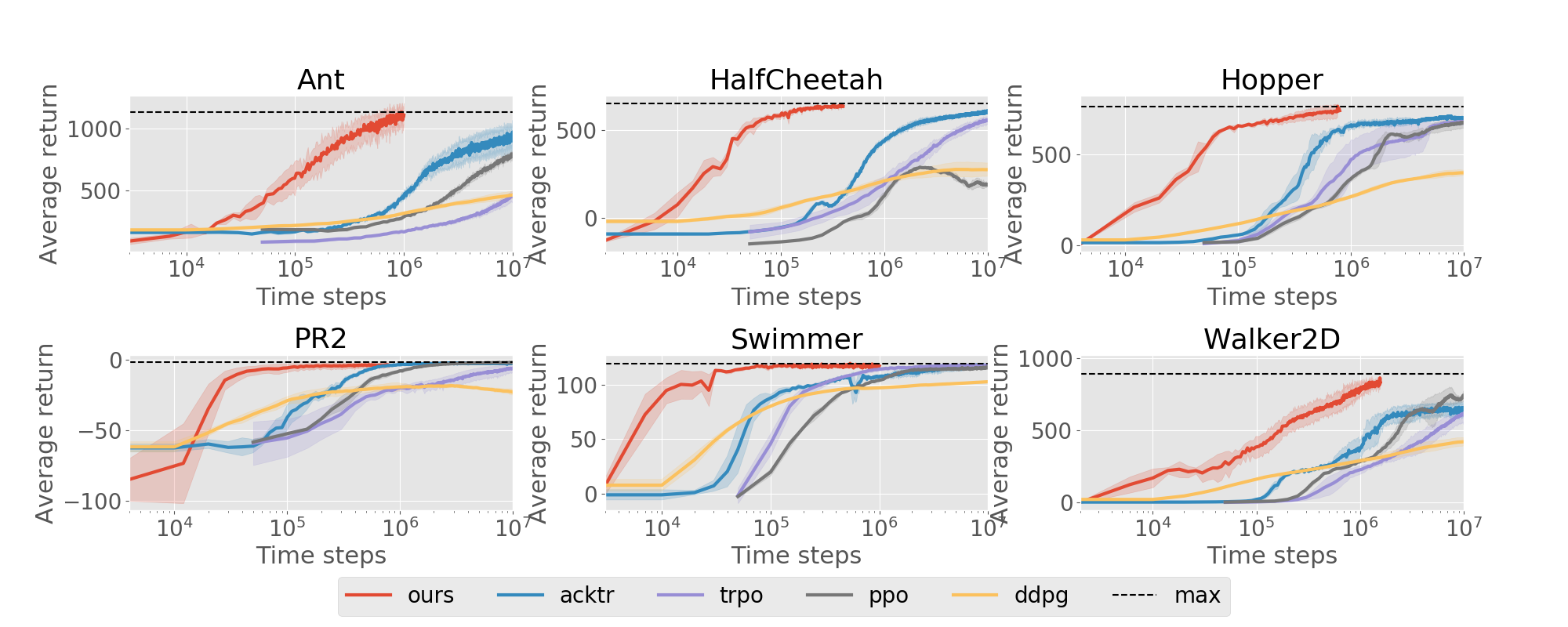}
\caption{Learning curves of \ours (``ours") and four state-of-the-art model-free methods in six different Mujoco environments with a horizon of 200.  \ours is able to match the asymptotic performance of model-free methods with two orders of magnitude less samples.}
\label{fig:mf}
\vspace{-0.1 in}
\end{figure}

We compare our method in sample complexity and performance to four state-of-the-art model free RL algorithms: Deep Deterministic Policy Gradient (DDPG)~\citep{lillicrap2015ddpg}, Trust Region Policy Optimization~\citep{Schulman2015}, Proximal Policy Optimization (PPO)~\citep{Schulman2017ProximalAlgorithms}, and Actor Critic using Kronecker-Factored Trust Region (ACKTR)~\citep{wu2017acktr}. The results are shown in Figure~\ref{fig:mf}.

In all the locomotion tasks we are able to achieve maximum performance using between 10 and 100 times less data than model-free methods. In the most challenging domains: ant, hopper, and walker2D; the data complexity of our method is two orders of magnitude less than the MF. In the easier tasks: the simulated PR2 and swimmer, our method achieves the same performance of MF using 20-50$\times$ less data. These results highlight the benefit of \ours for real robotics tasks; the amount of real-world data needed for attaining maximum return corresponds to 30 min in the case of easier domains and to 90 min in the more complex ones. 
\vspacesubsection
\subsection{Comparison to State-of-the-Art: Model-Based}
We also compare our method against recent model-based work: Model-Ensemble Trust-Region Policy Optimization (ME-TRPO)~\cite{Kurutach2018}, and the model-based approach introduced in ~\citet{Nagabandi2017a}, which uses MPC for planning (MB-MPC). 

\begin{figure}[h]
\vspace{-0.1 in}
\includegraphics[width=.95\textwidth]{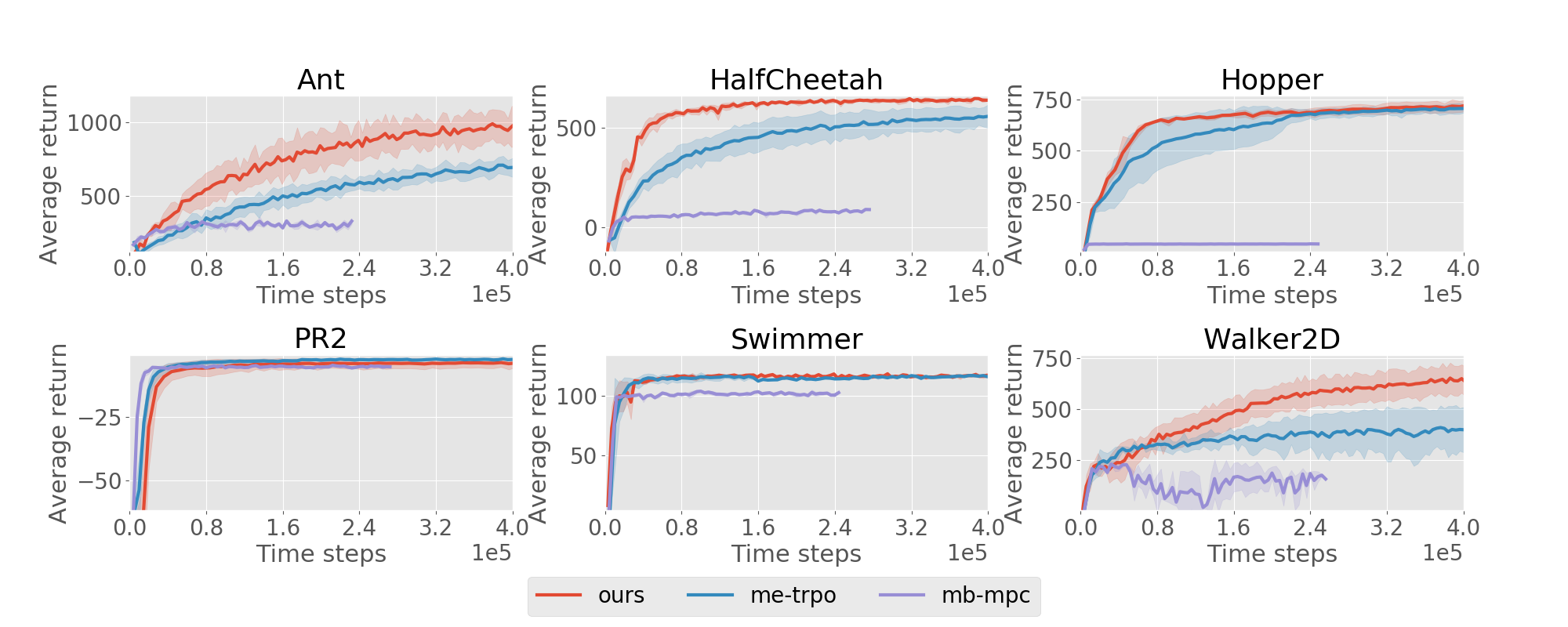}
\caption{Learning curves of \ours (``ours") and two MB methods in 6 different Mujoco environments with a horizon of 200.  \ours achieves better asymptotic performance and faster convergence rate than previous MB methods.}
\label{fig:mb}
\vspace{-0.1 in}
\end{figure}

\begin{wrapfigure}{r}{0.3\textwidth}
  \vspace{-0.2 in}
  \centering
  \includegraphics[width=.3\textwidth]{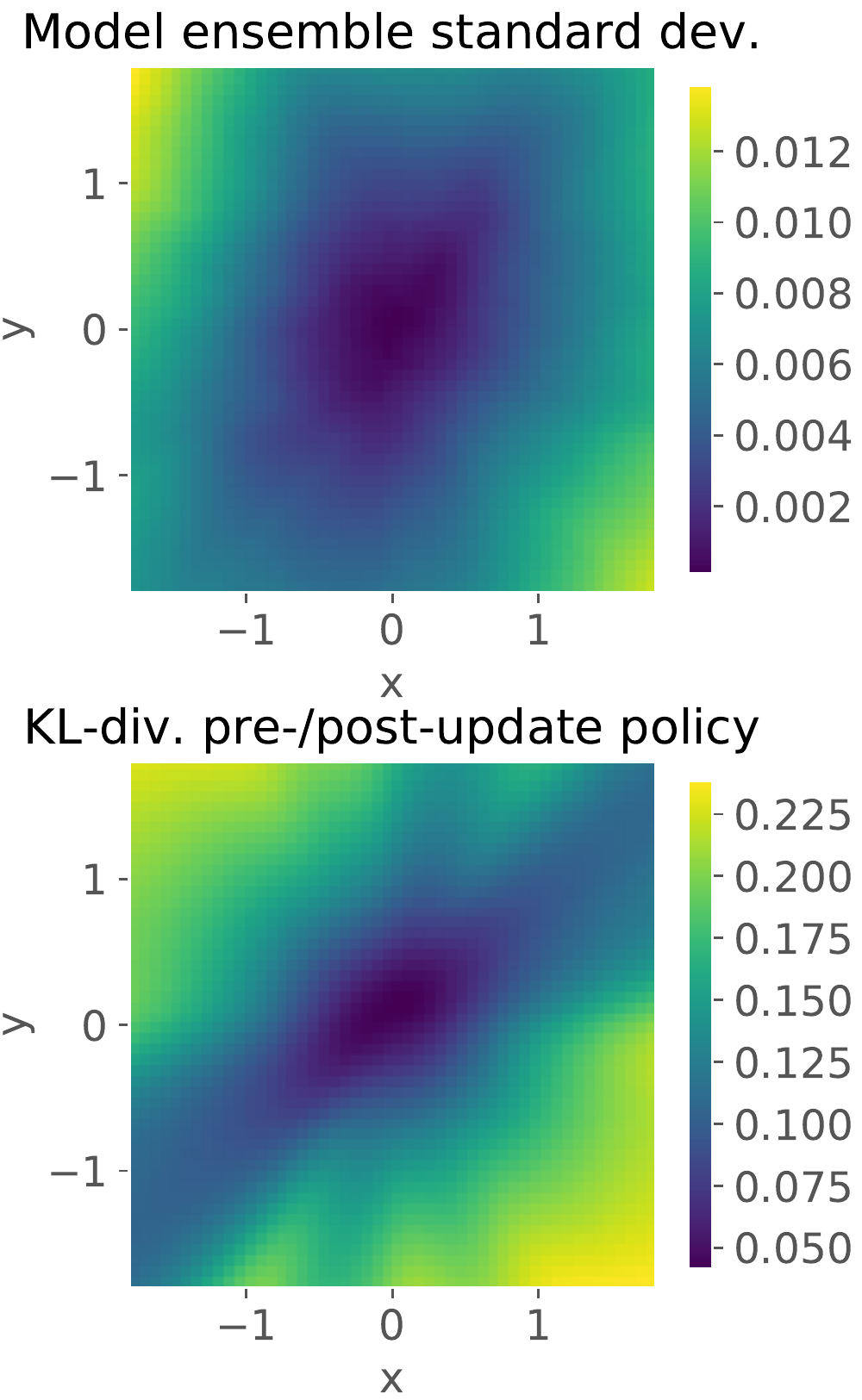}
  \vspace{-0.25 in}
  \caption{Upper: Standard deviation of model ensemble predictions Lower: KL-divergence between pre- and post-update policy (after 50 MB-MPO iterations in the 2-D Point env). The x and y axis denote the state-space dimensions of the 2-D Point environment}
    \label{figure:plasticity}
    \vspace{-0.5 in}
\end{wrapfigure}

The results, shown in Figure~\ref{fig:mb}, highlight the strength of \ours in complex tasks. MB-MPC struggles to perform well on tasks that require robust planning, and completely fails in tasks where medium/long-term planning is necessary (as in the case of hopper). In contrast, ME-TRPO is able to learn better policies, but the convergence to such policies is slower when compared to \ours. Furthermore, while ME-TRPO converges to suboptimal policies in complex domains, \ours is able to achieve max-performance.

\vspacesubsection
\subsection{Model Uncertainty and Policy Plasticity} 
\label{section:plasticity}

In section \ref{section:plasticity} we hypothesize that the meta-optimization steers the policy towards higher plasticity in regions with high dynamics model uncertainty while embedding consistent model predictions into the pre-update policy.
\added{To empirically analyze this hypothesis, we conduct an experiment in a simple 2D-Point environment where the agent, starting uniformly from $[-2,2]^2$, must go to the goal position $(0,0)$. We use the average KL-divergence
between $\pi_{\bm{\theta}}$ and the different adapted policies $\pi_{\bm{\theta'}_{k}}$ to measure the plasticity conditioned on the state $\bm{s}$.

Figure \ref{figure:plasticity} depicts the KL-divergence between the pre- and post-update policy, as well as the standard deviation of the predictions of the ensemble over the state space. Since the agent steers towards the center of the environment, more transition data is available in this region. As a result the models present higher accuracy in the center. The results indicate a strong positive correlation between model uncertainty and the KL-divergence between pre- and post-update policy. We find this connection between policy plasticity and predictive uncertainty consistently throughout the training and among different hyper-parameter configurations.
}

\vspacesubsection
\subsection{Robustness to Imperfect Dynamic Models and Compounding Errors} \label{exp:robustness}
We pose the question of how robust our proposed algorithm is w.r.t. imperfect dynamics predictions. We examine it in two ways. First, with an illustrative example of a model with clearly wrong dynamics. Specifically, we add biased Gaussian noise $\mathcal{N}(b,0.1^2)$ to the next state prediction, whereby the bias $b \sim \mathcal{U}(0,b_{\max})$ is re-sampled in every iteration for each model. Second, we present a realistic case on which long horizon predictions are needed. Bootstrapping the model predictions for long horizons leads to high compounding errors, making policy learning on such predictions challenging.

\begin{figure}[t]
\vspace{-0.1 in}
\includegraphics[width=.95\textwidth]{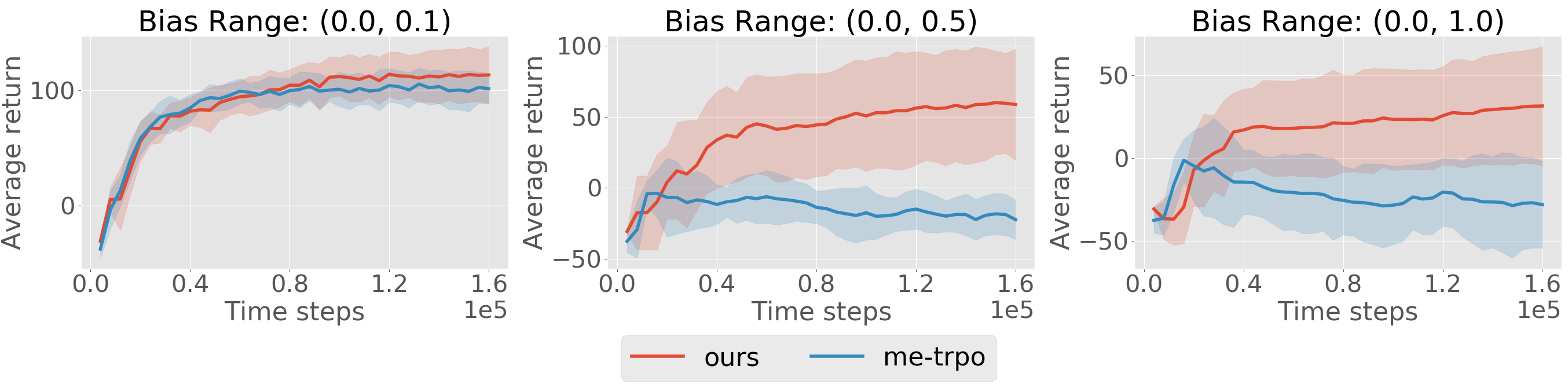}
\caption{Comparison of \ours (``ours") and ME-TRPO using 5 biased and noisy dynamic models in the half-cheetah environment with a horizon of 100 time steps. A bias term $b$ is sampled uniformly from a denoted interval in every iteration. During the iterations we add to the predicted observation a Gaussian noise $\mathcal{N}(b,0.1)$.}
\label{fig:robustness}
\vspace{-0.1 in}
\end{figure}

\begin{wrapfigure}{h}{0.4\textwidth}
\vspace{-0.2 in}
\centering
\includegraphics[width=0.4\textwidth]{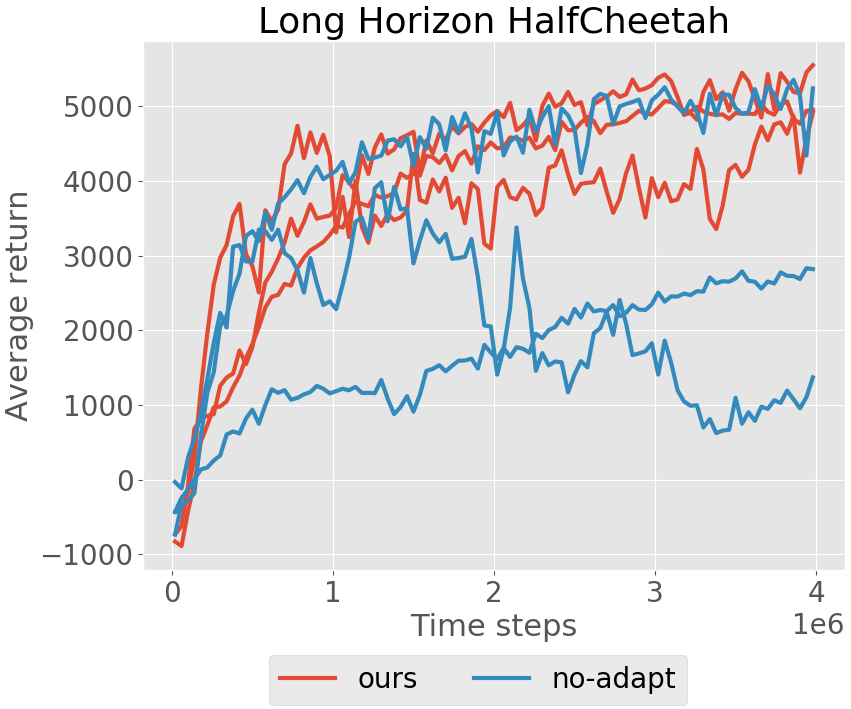}
\caption{Comparison of our method with and without adaptation. Depicted is the development of average returns during training with three different random seeds on the half-cheetah environment with a horizon of 1000 time steps.}
\label{fig:long_horizon}
\vspace{-0.2 in}
\end{wrapfigure}

Figure \ref{fig:robustness} depicts the performance comparison between our method and ME-TRPO on the half-cheetah environment for various values of $b_{\max}$. Results indicate that our method consistently outperforms ME-TRPO when exposed to biased and noisy dynamics models. ME-TPRO catastrophically fails to learn a policy in the presence of strong bias (i.e. $b_{\max}=0.5$ and $b_{\max}=1.0$), but our method, despite the strongly compromised dynamic predictions, is still able to learn a locomotion behavior with a positive forward velocity. 

This property also manifests itself in long horizon tasks. Figure~\ref{fig:long_horizon} compares the performance of our approach with inner learning rate $\alpha=10^{-3}$ against the edge case $\alpha=0$, where no adaption is taking place. For each random seed, \ours steadily converges to maximum performance. However, when there is no adaptation, the learning becomes unstable and different seeds exhibit different behavior: proper learning, getting stuck in sub-optimal behavior, and even unlearning good behaviors.

\vspacesection
\section{Conclusion}
\vspace{-0.05 in}
\label{sec:conclusion}
In this paper, we present a simple and generally applicable algorithm, model-based meta-policy optimization (MB-MPO), that learns an ensemble of dynamics models and meta-optimizes a policy for adaptation in each of the learned models. Our experimental results demonstrate that meta-learning a policy over an ensemble of learned models provides the recipe for reaching the same level of performance as state-of-the-art model-free methods with substantially lower sample complexity. We also compare our method against previous model-based approaches, obtaining better performance and faster convergence. Our analysis demonstrate the ineffectiveness of prior approaches to combat model-bias, and showcases the robustness of our method against imperfect models. As a result, we are able to extend model-based to more complex domains and longer horizons. One direction that merits further investigation is the usage of Bayesian neural networks, instead of ensembles, to learn a distribution of dynamics models. Finally, an exciting direction of future work is the application of \ours to real-world systems.

\clearpage
\acknowledgments{We thank A. Gupta, C. Finn, and T. Kurutach for the feedback on the earlier draft of the paper. IC was supported by La Caixa Fellowship. The research leading to these results received funding from the EU Horizon 2020 Research and Innovation programme under grant agreement No. 731761 (IMAGINE) and was supported by Berkeley Deep Drive, Amazon Web Services, and Huawei.}

{\small
\bibliography{references}
}

\clearpage
\appendix
\section{Appendix}
\subsection{Tailored Data Collection}
\label{appendix:taylored_data_colection}
We present the effects of collecting data using tailored exploration. We refer to tailored exploration as the effect of collecting data using the post-update policies -- the policies adapted to each specific model. When training policies on learned models they tend to exploit the deficiencies of the model, and thus overfitting to it. Using the post-update policies to collect data results in exploring the regions of the state space where these policies overfit and the model is inaccurate. Iteratively collecting data in the regions where the models are innacurate has been shown to greatly improve the performance~\cite{ross2011reduction}.

\begin{figure}[h]
    \centering
    \includegraphics[width=0.95\textwidth]{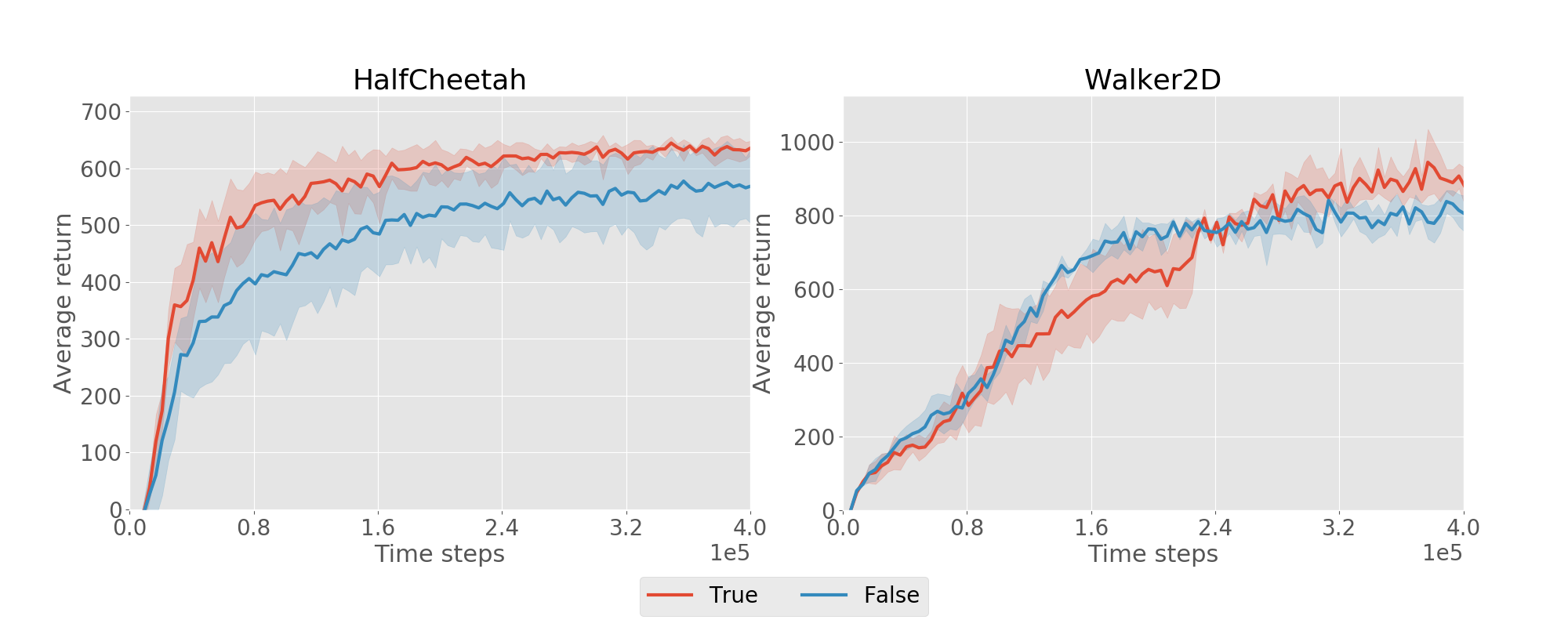}
    \caption{Tailored exploration study in the half-cheetah and walker2D environment. ``True" means the data is collected by using tailored exploration, and ``False" is the result of not using it, i.e., using the pre-update policy to collect data.}
    \label{fig:tailored_exploration}
\end{figure}

The effect of using tailored exploration is shown in Figure~\ref{fig:tailored_exploration}. In the half-cheetah and the walker we get an improvement of 12\% and 11\%, respectively. The tailored exploration effect cannot be accomplished by robust optimization algorithms, such as ME-TRPO. Those algorithms learn a single policy that is robust across models. The data collection using such policy will not exploit the regions in which each model fails resulting in less accurate models.

\subsection{Hyperparameter Study}
\label{exp:hyperparams}
\begin{figure}[h]
\includegraphics[width=.95\textwidth]{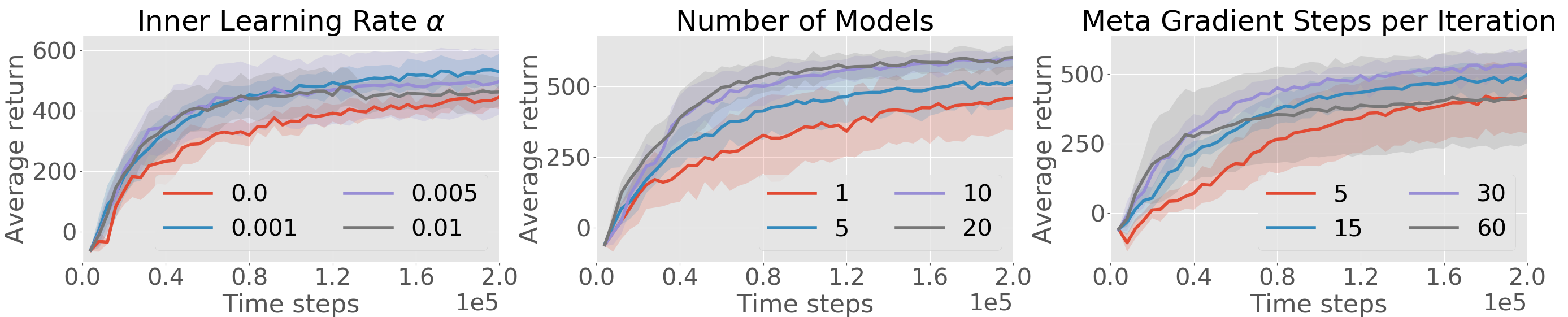}
\caption{Hyper-parameter study in the the half-cheetah environment of a) the inner learning rate $\alpha$, b) the number of dynamic models in the ensemble, and c) the number of meta gradient steps before collecting real environment samples and refitting the dynamic models.}
\label{fig:hyperparams}
\vspace{-0.1 in}
\end{figure}

We perform a hyperparameter study (see Figure~\ref{fig:hyperparams}) to assess the sensitivity of \ours to its parameters. Specifically, we vary the inner learning rate $\alpha$, the size of the ensemble, and the number of meta gradient steps before collecting further real environment samples. Consistent with the results in Figure~\ref{fig:long_horizon}, we find that adaptation significantly improves the performance when compared to the non-adaptive case of $\alpha=0$. Increasing the number of models and meta gradient steps per iteration results in higher performance at a computational cost. However, as the computational burden is increased the performance gains diminish. 

Up to a certain level, increasing the number of meta gradient steps per iteration improves performance. Though, too many meta gradients steps (i.e. 60) can lead to early convergence to a suboptimal policy. This may be due to the fact that the variance of the Gaussian policy distribution is also learned. Usually, the policies variance decreases during the training. If the number of meta-gradient steps is too large, the policy loses its exploration capabilities too early and can hardly improve once the models are more accurate. This problem can be alleviated using a fixed policy variance, or by adding an entropy bonus the learning objective.

\subsection{Experiment Setup} \label{appendix:experiment_setup}
In the following we provide a detailed description of the setup used in the experiments presented in section \ref{sec:experiments}:

\textbf{Environments:}
\begin{figure}[h]
\centering
\includegraphics[width=.15\textwidth]{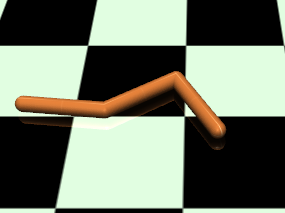}
\includegraphics[width=.15\textwidth]{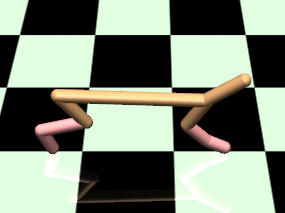}
\includegraphics[width=.15\textwidth]{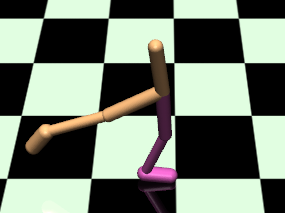}
\includegraphics[width=.15\textwidth]{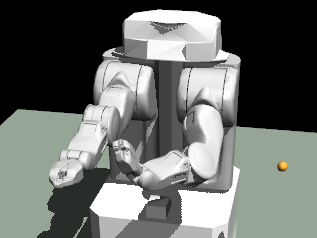}
\includegraphics[width=.15\textwidth]{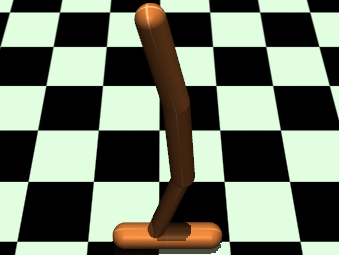}
\includegraphics[width=.15\textwidth]{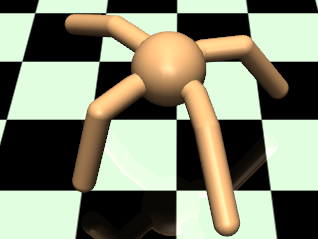}
\caption{Mujoco environments used in our experiments. Form left to right: swimmer, half-cheetah, walker2D, PR2, hopper, and ant.}
\label{fig:mujoco-envs}
\vspace{-0.15 in}
\end{figure}

We benchmark MB-MPO on six continuous control benchmark tasks in the Mujoco simulator \cite{mujoco}, shown in Fig.~\ref{fig:mujoco-envs}. Five of these tasks, namely swimmer, half-cheetah, walker2D, hopper and ant, involve robotic locomotion and are provided trough the OpenAI gym~\cite{openai-gym}. 

The sixth, the 7-DoF arm of the PR2 robot, has to reach arbitrary end-effector positions. Thereby, the PR2 robot is torque controlled. The reward function is comprised of the squared distance of the end-effector (TCP) to the goal and energy / control costs:

\begin{equation*}
    r(s,a) = - ||s_{\text{TCP}} - x_{\text{goal}}||^2_2 - 0.05 * ||a||^2_2
\end{equation*}

In section \ref{section:plasticity} we use the simple 2D-Point environment to analyze the connection between policy plasticity and model uncertainty. The corresponding MDP is defined as follows:
\begin{align*}
    \mathcal{S} &= \mathbb{R}^2 \\
    \mathcal{A} &= [-0.1, 0.1]^2 \\
    p_0(s_0) &= \mathcal{U}_{[-2, 2]^2}(s_0) ~~ \text{(uniform distribution over } [-2, 2]^2 ) \\
    p(s_{t+1}|s_t, a_t) &= \delta(s_t + a_t) \\ 
    r(s_t, a_t) &= - || s_{t} ||_2^2 \\
    H &= 30 
\end{align*}

\textbf{Policy:} We use a Gaussian policy $\pi_\theta(a|s) = \mathcal{N}(a | \mu(a)_{\theta_\mu}, \sigma_{\theta_\sigma}$) with diagonal covariance matrix. The mean $\mu(a)_{\theta_\mu}$ is computed by a neural network (2 hidden layers of size 32, tanh nonlinearity) which receives the current state $s$ as an input. During the policy optimization, both the weights $\theta_\mu$ of the neural network and the standard deviation vector $\sigma_{\theta_\sigma}$ are learned.

\textbf{Advantage-Estimation:} We use generalized advantage estimation (GAE) \citep{Schulmanetal_ICLR2016} with $\gamma=0.99$ and $\lambda=1$ in conjunction with a linear reward baseline as in \citep{rllab} to estimate advantages.

\textbf{Dynamics Model Ensemble:} In all experiments (except in Figure \ref{fig:hyperparams}b) we use an ensemble of 5 fully connected neural networks. For the different environments the following hidden layer sizes were used:
\begin{itemize}
    \item Ant, Walker: (512, 512, 512)
    \item PR2, Swimmer, Hopper, Half-Cheetah: (512, 512)
    \item 2D-Point-Env: (128, 128)
\end{itemize}
In all models, we used weight normalization and ReLu nonlinearities. For the minimization of the $l^2$ prediction error, the Adam optimizer with a batch-size of 500 was employed. In the first iteration all models are randomly initialized. In later iterations, the models are trained with warm starts using the parameters of the previous iteration. In each iteration and for each model in the ensemble the transition data buffer $\mathcal{D}$ is randomly split in a training (80\%) and validation (20\%) set. The latter split is used to compute the validation loss after each training epoch on the shuffled training split. A rolling average of the validation losses with a persistence of 0.95 is maintained throughout the epochs. Each model's training is stopped individually as soon as the rolling validation loss average decreases.

\textbf{Meta-Policy Optimization:} As described in section \ref{sec:meta-training}, the policy parameters $\theta$ are optimized using the gradient-based meta learning framework MAML. For the inner adaptation step we use a gradient step-size of $\alpha = 0.001$. For maximizing the meta-objective specified in equation \ref{eq:meta-objective} we use the policy gradient method TPRO \citep{Schulman2015} with KL-constraint $\delta=0.01$. Since computing the gradients of the meta-objective involves second order terms such as the Hessian of the policy's log-likelihood, computing the necessary Hessian vector products for TRPO analytically is very compute intensive. Hence, we use a finite difference approximation of the vector product of the Fisher Information Matrix and the gradients as suggested in \cite{Finn2017}. If not denoted differently, 30 meta-optimization steps are performed before new trajectories are collected from the real environment.

\textbf{Trajectory collection: } In each algorithm iteration 4000 environment transitions (20 trajectories of 200 time steps) are collected. For the meta-optimization, 100000 imaginary environment transitions are sampled.

\subsection{Computational Analysis}
In this section we compare the computational complexity of \ours against TRPO. Specifically, we report the wall clock time that it takes both algorithms to reach maximum performance on the half-cheetah environment when running the experiments on an Amazon Web Services EC2 c4.4xlarge compute instance. Our method only requires 20\% more compute time than TRPO (7 hours instead of 5.5), while attaining 70$\times$ reduction in sample complexity. The main time bottleneck of our method compared with the model-free algorithms is training the models.

Notice that when running real world experiment, our method will be significantly faster than model-free approaches since the bottleneck then would shift towards the data collection step.

\end{document}